\documentclass{article}%
\usepackage{worldmind_preprint,times}

\usepackage{amsmath,amsfonts,bm}

\def\eqref#1{equation~\ref{#1}}

\def\1{\bm{1}}

\DeclareMathAlphabet{\mathsfit}{\encodingdefault}{\sfdefault}{m}{sl}
\SetMathAlphabet{\mathsfit}{bold}{\encodingdefault}{\sfdefault}{bx}{n}

\usepackage{hyperref}
\usepackage{url}
\usepackage{amssymb}
\usepackage{graphicx}
\usepackage{wrapfig}
\usepackage{needspace}
\usepackage{booktabs}
\usepackage{multirow}
\usepackage{subcaption}
\title{WorldMind: Test-Time Scaling for\\
Visual Spatial Reasoning through\\
Latent World Simulation}

\author{%
  Haoming Wang$^{1}$\quad Yongjian Fu$^{2}$\quad Ju Ren$^{2}$\quad Wei Gao$^{1}$\\[4pt]
  $^{1}$University of Pittsburgh\\
  $^{2}$Tsinghua University
}
\date{}
\hypersetup{
  pdftitle={WorldMind: Test-Time Scaling for Visual Spatial Reasoning through Latent World Simulation},
  pdfauthor={Haoming Wang, Yongjian Fu, Ju Ren, Wei Gao},
  hidelinks
}
\begin{document}

\maketitle

\begin{abstract}
Vision-Language Models (VLMs) have continuous deficiency in spatial reasoning tasks that require \emph{mental spatial transformation}, such as reasoning under hypothetical changes in viewpoint or position, known as perspective-taking tasks. Existing approaches either scale linguistic reasoning or augment VLMs with static geometric representations, but neither directly enables active transformation of the represented scene. World models offer this capability, yet existing systems typically communicate through synthesized RGB images, introducing a latent-to-pixel-to-latent bottleneck. We introduce \emph{WorldMind}, a framework for test-time scaling of spatial reasoning through latent world simulation. Given a query and observation, the VLM selects useful spatial transformations, while a world model executes them and exposes aligned intermediate latent features directly to the VLM. The resulting latent evidence is incorporated into subsequent reasoning steps, enabling iterative spatial simulation without repeatedly decoding imagined states into pixels. To optimize this interaction, WorldMind further employs process-verifiable reinforcement learning that rewards reasoning progress only when the associated action and latent transition are valid. WorldMind thus scales spatial reasoning by allocating additional inference-time computation to active world simulation rather than generating more language or pixels. Across VSI-Bench, MMSI-Bench, and MindCube-Tiny, WorldMind achieves an average improvement of 16.0\% over the base VLM and 8.3\% over the strongest RGB-based world simulation.
\end{abstract}

\begin{wrapfigure}{r}{2.3in}
	\centering
	\vspace{-0.3in}
	\includegraphics[width=2.3in]{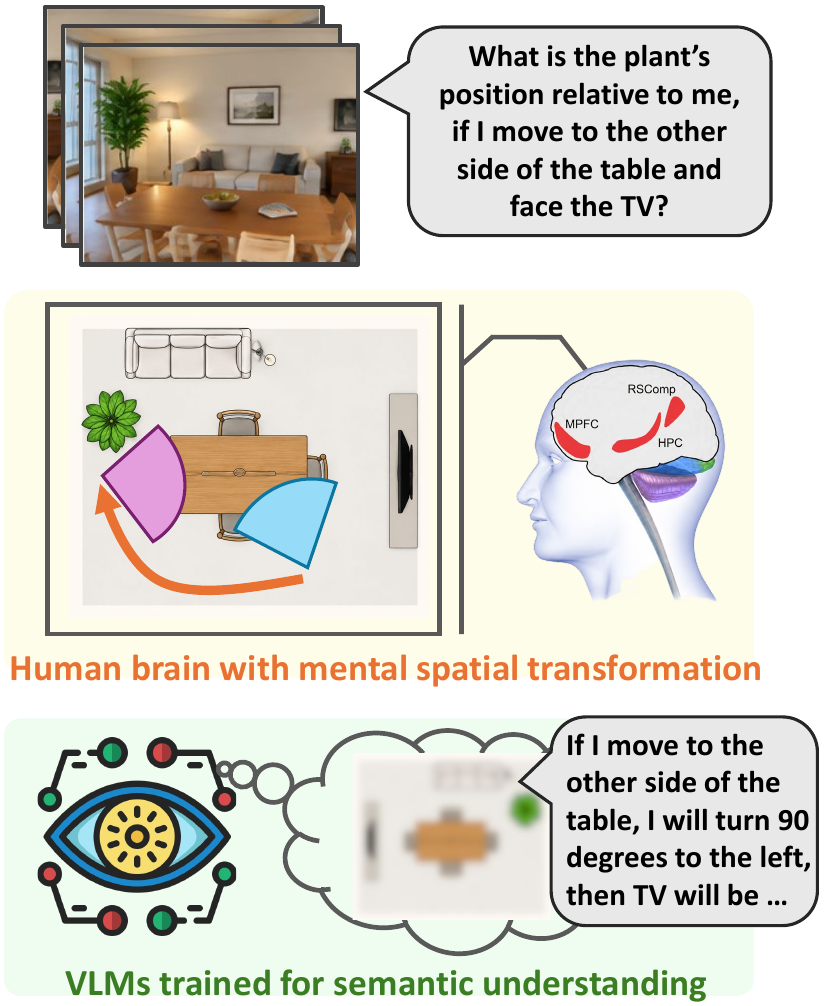}
	\vspace{-0.25in}
	\caption{Spatial transformation in perspective-taking reasoning tasks}
	\vspace{-0.15in}
	\label{fig:motivation}
\end{wrapfigure}

\section{Introduction}

Vision-Language Models (VLMs) have achieved remarkable success in multimodal perception, and demonstrated strong capabilities in semantic recognition and extraction of  basic spatial cues from individual frames [\citenum{chen2024spatialvlm,liu2024improved,deitke2025molmo}]. However, their performance degrades substantially when spatial reasoning requires going beyond what is directly observed, as shown in Figure \ref{fig:motivation}, such as determining how objects would appear from another viewpoint, mentally rotating a scene, or planning a sequence of viewpoint changes [\citenum{liu2025spatial,xue2026reasoning,cai2026scaling,ma2022sqa3d}]. From the perspective of cognitive science, the limitation of VLMs in these perspective-taking tasks is due to their intrinsic deficiency in reproducing the human's spatial intelligence, which builds on \emph{mental spatial transformation}, the ability to construct an internal representation of the environment and transform that representation under hypothetical changes in viewpoint, orientation, or position [\citenum{shepard1971mental,hegarty2004dissociation,huttenlocher1979coding}]. Current VLMs, since being trained for semantic understanding, lack a mechanism for similarly performing such internal spatial transformation, and hence struggle on complex downstream tasks that require reasoning over unobserved or transformed states, including multi-step navigation, dynamic perspective taking, and reasoning about occluded geometry [\citenum{yang2025thinking,zhang2025call,chen2025spatial}].

\begin{wrapfigure}{r}{2.8in}
	\centering
	\vspace{-0.2in}
	\includegraphics[width=2.8in]{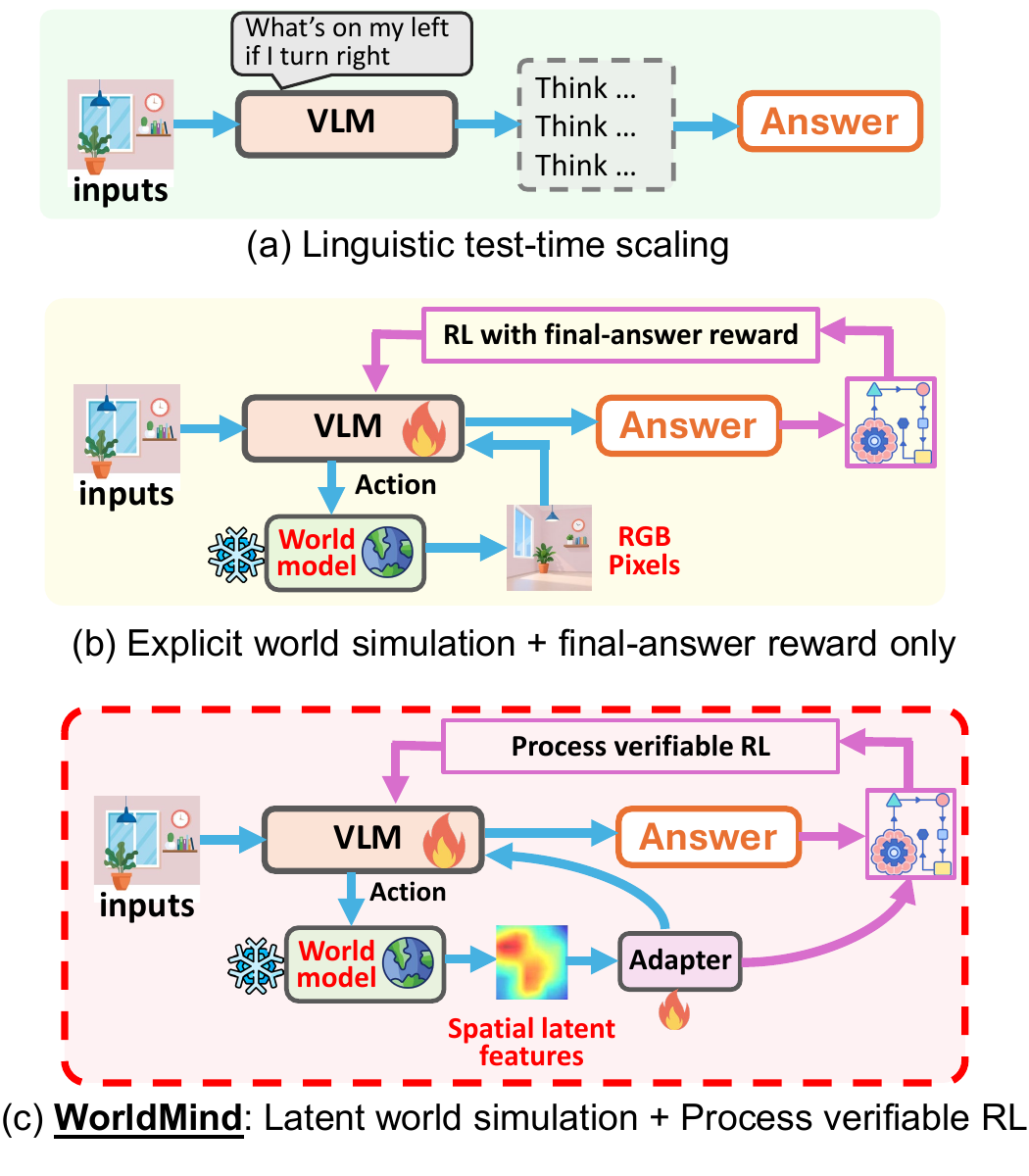}
	\vspace{-0.3in}
	\caption{Comparisons of test-time scaling paradigms for spatial reasoning}
	\vspace{-0.15in}
	\label{fig:comparison}
\end{wrapfigure}

Existing efforts on addressing this gap primarily follow two approaches but both have fundamental limitations. The first approach is to scale linguistic test-time computation for mental spatial transformation, as shown in Figure \ref{fig:comparison}(a). This strategy is highly effective in symbolic and mathematical reasoning [\citenum{wei2022chain,snell2025scaling}], but simply generating more texts at test time often fails to improve or even degrade spatial reasoning, because the missing spatial information about the real-world scene cannot be made available by extra linguistic texts, which may instead introduce extra bias or confusions [\citenum{yang2025thinking,kancheti2026chain,ji2025enhancing}]. The second approach is to inject encoded 3D representations (e.g. depth maps or point clouds) into the VLMs' inputs [\citenum{cai2025spatialbot,liang2026enhancing,cheng20263d}]. While these representations provide useful geometric cues for the currently observed scene and benefit localized metric tasks such as object measurement, they remain limited to the existing observations and cannot support tasks that require active viewpoint transformation or exploration, such as reasoning from an opposing observer's perspective [\citenum{yin2025spatial,kim2026diagnosing}].

Recently emerged models for novel view synthesis, such as world action models [\citenum{zhang2026worldstereo,zhou2025stable,ren2025gen3c}], are naturally a better alternative to enable spatial transformations and provide spatial information beyond current observations, because they are trained to predict how visual observations change under camera motions or explicitly specified actions. As a result, their learning objectives inherently require modeling how the environment changes under hypothetical transformations of viewpoint and state. More importantly, recent analyses suggest that their intermediate latent representations encode rich information about scene geometry, camera motion, object correspondence, and spatial structures, even without explicit 3D supervision [\citenum{huang2026much,zhang2025can,man2024lexicon3d}]. This means that these models serve as \textbf{spatial simulators}, whose internal representations provide structured spatial information that can be directly exploited for spatial reasoning.

Some existing approaches have coupled VLMs with world models for active visual imagination [\citenum{zhang2026think3d,zhu2026thinking,zhang2026thinking}]. Given a spatial query, the VLM selects a target viewpoint, and the world model generates the novel view being needed as additional frames for reasoning. These methods establish an important paradigm of test-time scaling for spatial reasoning, through actively imagined observations in new viewpoints. However, they use the world model primarily as an \emph{\textbf{explicit} image renderer} [\citenum{cao2026spatialdreamer,yang2026mindjourney}]: the world model first converts its internal spatial representation into explicit pixels, from which the VLM recovers the spatial structures through its visual encoder, as shown in Figure \ref{fig:comparison}(b). Since the visual encoder used in VLMs (e.g., ViT [\citenum{dosovitskiy2020image}]) are mostly optimized for visual-semantic understanding, such recovery in VLMs would certainly cause loss or ineffective representation of important spatial information, such as geometry, object correspondence and relationship among objects.

In this paper, we present \emph{WorldMind}, a new framework for test-time scaling of spatial reasoning through \textbf{latent world simulation}. As shown in Figure \ref{fig:comparison}(c), given a spatial query and the current observation, the VLM iteratively decides \emph{what to simulate} by selecting a useful viewpoint transformation, and the world model performs the scene transformation and returns its \emph{latent representation} to VLM as spatial features. In this way, rather than relying on the synthesized pixels, WorldMind directly extracts the spatially informative latent representations from the world model, aligns them with the VLM, and incorporates them into VLM's reasoning. This interaction continues until sufficient evidence has been gathered to answer the query.%

Realizing this latent world simulation poses two primary technical challenges. The first challenge is to determine which representations of the world model should be exposed to the VLM. The internal representations of diffusion world models are distributed across multiple network blocks and denoising steps, and features at different locations encode different types of information, ranging from low-level appearance and texture to object semantics, geometry and global scene structure [\citenum{wang2026demystifying}].%
WorldMind identifies spatially informative latent representations by evaluating features across diffusion layers and denoising steps, retaining those most useful for spatial question answering, and aligns them with the VLM through a lightweight adapter trained with answer supervision (Section \ref{sec:alignment}).

The second challenge is to effectively learn multi-step latent simulation. Final-answer supervision alone provides little guidance on whether each intermediate simulation step is useful or correctly executed [\citenum{uesato2022solving}]. To address this challenge, WorldMind trains the simulation policy in two stages: supervised fine-tuning on synthetic reasoning trajectories to bootstrap viewpoint selection (Section \ref{sec:sft}), followed by reinforcement learning with step-level rewards that encourage task progress, correct execution of spatial transformations, and reliable simulation (Section \ref{sec:rl}).

To the best of our knowledge, WorldMind is the first that scales spatial reasoning at test time by directly exposing the world model's latent spatial representations to a VLM. It integrates latent feature selection, alignment, and multi-step world simulation into a unified reasoning framework, so that the VLM can actively acquire, interpret, and directly reason over transformed spatial states without repeatedly decoding them into pixels. We evaluated WorldMind across multiple spatial reasoning benchmarks, and experiment results show that WorldMind improves the average success rate of spatial reasoning tasks by 16.0\% over the base VLMs and 8.3\% over explicit world simulation, with particularly strong gains on viewpoint transformation and multi-step spatial reasoning tasks.

\section{Motivation and Related Work}

\begin{wrapfigure}{r}{2.4in}
    \centering
    \vspace{-0.20in}

    \begin{subfigure}{\linewidth}
        \centering
        \includegraphics[width=2.35in]{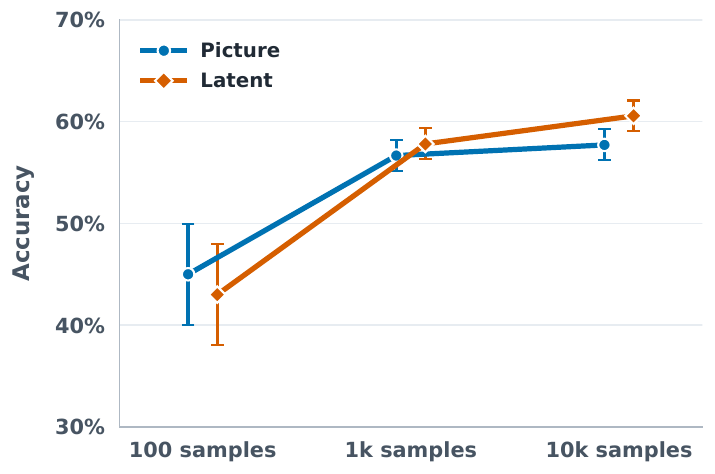}
        \caption{Performance under different fine-tuning data scales.}
        \label{fig:prelim_scale}
    \end{subfigure}

    \vspace{1mm}

    \begin{subfigure}{\linewidth}
        \centering
        \scriptsize
        \setlength{\tabcolsep}{2.8pt}
        \renewcommand{\arraystretch}{1.02}
        \begin{tabular}{lcccc}
            \toprule
            \textbf{Input}
            & \textbf{obj.cnt.}
            & \textbf{rel.dir}
            & \textbf{route}
            & \textbf{app.order} \\
            \midrule
            RGB
            & 68.4 & 65.2 & 35.8 & 69.7 \\
            WM Latent
            & \textbf{69.3}
            & \textbf{67.2}
            & \textbf{38.1}
            & \textbf{69.4} \\
            \bottomrule
        \end{tabular}
        \caption{Task-wise comparison on VSI-bench.}
        \label{fig:prelim_task}
    \end{subfigure}

    \vspace{-1mm}
    \caption{\textbf{Preliminary comparison between RGB and world-model latent representations.}
    (a) Comparison under different fine-tuning data scales.
    (b) Comparison across VSI-bench tasks. }
    \label{fig:prelim}
    \vspace{-0.15in}
\end{wrapfigure}

Recent studies show that intermediate representations of video diffusion and world models encode substantial 3D and dynamic information [\citenum{huang2026much,zhang2025can}]. These observations motivate our design of WorldMind to directly expose the selected latent representations of world model to the VLM, avoiding repeated decoding and re-encoding of explicit pixels as the existing approaches do [\citenum{yang2026mindjourney,cao2026spatialdreamer,zhu2026thinking,zhang2026thinking}], while retaining active, query-driven world simulation at test time.

To examine whether the world model's latent states are essentially better spatial representations compared to explicit pixels in VLM's spatial reasoning, we conduct a preliminary study using Qwen3-VL-8B-Instruct [\citenum{bai2025qwen3vltechnicalreport}] on the MindCube benchmark [\citenum{yin2025spatial}], by comparing the model fine-tuned  with RGB inputs against the model fine-tuned with world-model latent features under different training-data scales. For this preliminary study, we uniformly sample latent features over layers and denoise steps. As shown in Figure ~\ref{fig:prelim}, RGB inputs perform better with small data scale, because the pretrained VLM is already well aligned with the image modality. As the amount of fine-tuning data increases, the latent-based model catches up with and eventually surpasses the RGB-based counterpart, suggesting that world-model latents can provide more effective spatial representations once sufficient alignment data are available. We further break down the comparison across different spatial reasoning tasks on VSI-bench \citenum{yang2025thinking}. The advantage of latent representations is most pronounced on tasks requiring explicit transformation of the observed spatial state, such as relative direction (rel.dir) and route plan (route).

\begin{figure}[h]
	\centering
	\vspace{-0.15in}
	\includegraphics[width=0.85\linewidth]{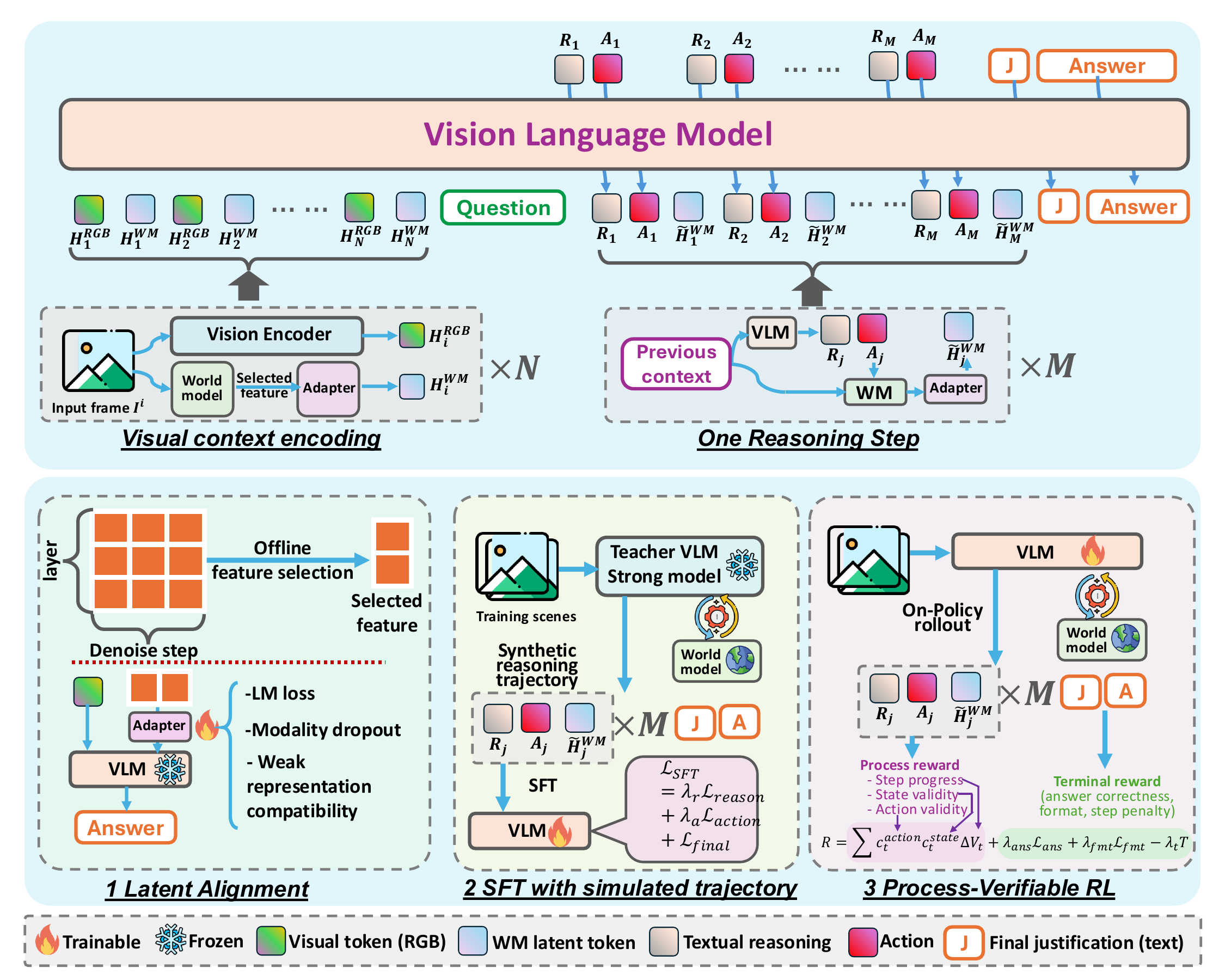}
	\vspace{-0.05in}
	\caption{Overview of WorldMind design. Given the currently observed RGB frames and a spatial query, WorldMind augments standard VLM visual tokens with the selected latent features from a pre-trained world model, and performs iterative reasoning through camera actions and latent world simulation. To ensure proper latent world simulation, WorldMind first trains an adapter that selects latent features from the world model and aligns them with a frozen VLM. Then, it fine-tunes the VLM in two stages: (1) supervised fine-tuning with simulated reasoning trajectories; and (2) process-verifiable reinforcement learning, which further optimizes action selection and reasoning using intermediate and final rewards.}
	\vspace{-0.2in}
	\label{fig:overview}
\end{figure}

\section{Method}
\label{sec:method}

As shown in Figure \ref{fig:overview}, given a sequence of current RGB observations $\{I_1,\ldots,I_N\}$ and a question $Q$, WorldMind augments the standard VLM visual representation with latent features extracted from a pre-trained world model. For each frame, the VLM produces $H_i^{\mathrm{rgb}}$ that encodes the observed RGB frame $I_i$ in the VLM's native visual representation space, and the corresponding world-model latent $Z_i$ is mapped into the VLM feature space through a learnable adapter as $H_i^w=P_w(Z_i)$. The initial context is constructed as $X_0=[H_1^{\mathrm{rgb}},H_1^w,\ldots,H_N^{\mathrm{rgb}},H_N^w,Q]$.

During reasoning, the VLM iteratively generates a textual reasoning token $R_t$ and a camera action $A_t$. This action is executed by the world model, which returns a new latent state $\widetilde{H}_t^w$ that is appended to the VLM's reasoning context. The complete inference sequence is $X_0 \rightarrow [R_1,A_1,\widetilde{H}_1^w] \rightarrow \cdots \rightarrow [R_T,A_T,\widetilde{H}_T^w] \rightarrow [J,Y]$, where $J$ and $Y$ denote the final justification and answer. In this way, additional test-time computation is used to acquire new latent spatial evidence rather than merely generating more texts.

The key to ensuring proper latent world simulation described above is to select spatially informative world-model representations, align them with the VLM's feature space, and train the VLM to actively query and utilize these latent states during multi-step reasoning. To achieve this, we first train an adapter for \textbf{latent feature selection and alignment}, to identify a compact set of spatially informative world model representations and align them with a frozen VLM's representation space. Afterwards, it fine-tunes the VLM being used for spatial reasoning in two stages: (1) \textbf{simulation-trajectory supervised fine-tuning} trains the adapter and VLM's LoRA modules on synthetic trajectories, to interleave reasoning, camera-action selection, world-model simulation, and latent-state conditioning; and (2) \textbf{process-verifiable reinforcement learning} further optimizes the policy on its own sampled trajectories, rewarding the reasoning progress only when the intended spatial transformation is successfully realized and the resulting latent state is reliable, while retaining terminal answer supervision and penalizing unnecessary world-model calls. Together, these stages progressively teach WorldMind to interpret, reason with, and actively acquire latent spatial evidence.

\Needspace{18\baselineskip}
\subsection{Latent Selection and Alignment}
\label{sec:alignment}

\begin{wrapfigure}{r}{2.3in}
	\centering
	\includegraphics[width=\linewidth]{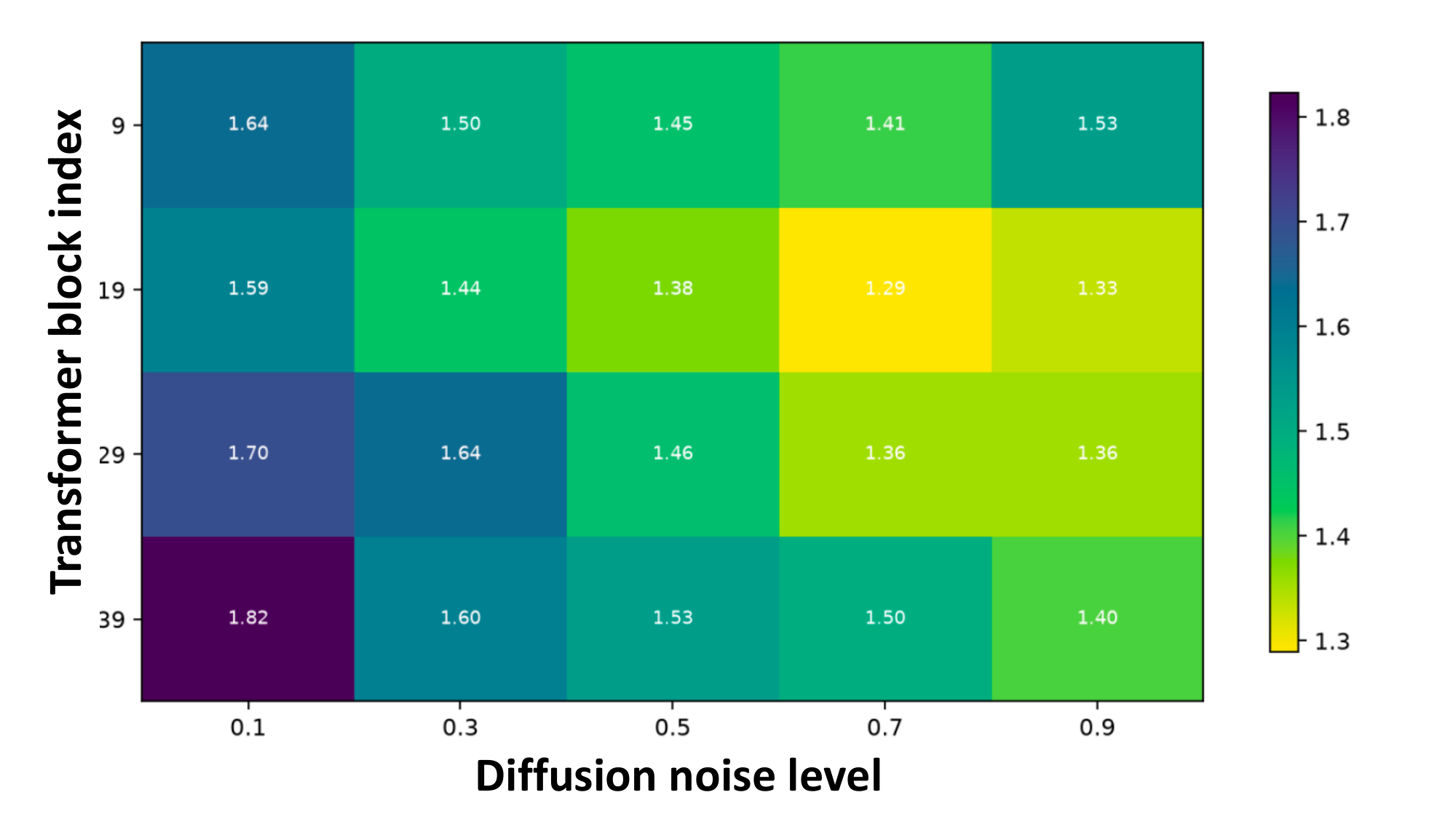}
	\captionsetup{justification=raggedright,singlelinecheck=false}
	\caption{Spatial information across world-model latents.}
	\label{fig:wm_probe}
\end{wrapfigure}

World-model representations vary substantially across network layers and denoising timesteps, and not all latent features are equally informative for spatial reasoning. To show this empirically, we conduct a simple probing experiment in which a lightweight MLP is trained to reconstruct the scene's 3D point cloud from latent features extracted at different transformer blocks and diffusion noise levels. As shown in Fig.~\ref{fig:wm_probe}, reconstruction error varies considerably across layer--timestep combinations, with some intermediate representations preserving substantially more geometric information than others. This observation motivates us to avoid arbitrarily selecting a single latent feature. Instead, we first identify a compact subset of spatially informative representations and then align the selected features with the pretrained VLM for downstream reasoning.

\vspace{-0.1in}
\paragraph{Learned latent selection.}
We construct a candidate feature bank $\mathcal{F}=\{h^{(\ell,\tau)}\}$ from multiple diffusion layers $\ell$ and denoising timesteps $\tau$, and train a temporary fusion adapter over the full set. We then estimate the contribution of each layer--timestep pair using attribution or feature ablation and retain the top-$K$ features, forming a fixed subset $\mathcal{S}^{*}=\{(\ell_1,\tau_1),\ldots,(\ell_K,\tau_K)\}$. This selection is performed offline and kept fixed during subsequent training and inference.

Using the selected features, we learn a lightweight adapter $\mathcal{P}_w$ that maps world-model representations into a form usable by the pretrained VLM. For frame $I_i$, the resulting latent representation is $H_i^w=\mathcal{P}_w(\{h_i^{(\ell,\tau)}:(\ell,\tau)\in\mathcal{S}^{*}\})$. During alignment, the world model and VLM backbone remain frozen, while only $\mathcal{P}_w$ is optimized. Rather than enforcing token-level matching with RGB features, we perform \emph{functional alignment}, allowing the projected latents to preserve complementary geometric information such as depth, viewpoint, and 3D structure. Given the target response $Y^{\mathrm{align}}=[J,Y]$, we optimize the standard autoregressive objective $\mathcal{L}_{\mathrm{ans}}=-\sum_k \log p_\theta(y_k\mid X_{\mathrm{align}},y_{<k})$.

\vspace{-0.1in}
\paragraph{Modality dropout.}
Since $I_i$ and $Z_i$ describe the same scene, the frozen VLM may rely on RGB alone and ignore the world-model tokens. To mitigate this modality dominance [\citenum{piekenbrinck2024rgb}], we randomly train with either the joint input $X_{\mathrm{joint}}=[H^{\mathrm{rgb}},H^w,Q]$ or a latent-only input $X_{\mathrm{latent}}=[H^w,Q]$, sampled with probabilities $1-p_{\mathrm{drop}}$ and $p_{\mathrm{drop}}$, respectively. The latent-only samples force the latent interface to expose task-relevant information from the world-model representation.

\vspace{-0.1in}
\paragraph{Weak representation compatibility.}
We further introduce an optional global compatibility objective that encourages corresponding RGB and world-model representations to occupy a similar semantic space without enforcing token-level matching. We compute pooled projections $c_i^{w}=g_w(\operatorname{Pool}(H_i^{w}))$ and $c_i^{\mathrm{rgb}}=\operatorname{sg}(g_v(\operatorname{Pool}(H_i^{\mathrm{rgb}})))$, normalize them, and apply a contrastive loss $\mathcal{L}_{\mathrm{compat}}$ over matched and mismatched pairs within the batch [\citenum{girdhar2023imagebind,xue2023ulip}]. The overall alignment objective is
\begin{equation}
	\mathcal{L}_{\mathrm{align}}
	=
	\mathcal{L}_{\mathrm{ans}}
	+
	\lambda_{\mathrm{compat}}
	\mathcal{L}_{\mathrm{compat}},
	\label{eq:alignment_objective}
\end{equation}
where $\lambda_{\mathrm{compat}}$ is kept small so that functional answer supervision remains dominant. The compatibility branch is used only during training and discarded at inference.

\subsection{Simulation-Trajectory Supervised Fine-Tuning}
\label{sec:sft}
After latent alignment, the model can interpret world-model features but has not yet learned how to use them within an iterative reasoning process.  This stage teaches the VLM to interact with the world model by identifying missing spatial information, selecting useful camera transformations, and incorporating the resulting latent states into its reasoning.

Such training requires step-level supervision over both textual reasoning and camera actions. Given an initial context, we assume a target trajectory of the form $[R_1^{*},A_1^{*},\widetilde{H}_1^{w},\ldots,R_T^{*},A_T^{*},\widetilde{H}_T^{w},J^{*},Y^{*}]$, where $R_t^{*}$ and $A_t^{*}$ denote the target reasoning step and camera action, and $\widetilde{H}_t^{w}$ is the world-model latent returned after executing $A_t^{*}$. The latent states are treated as observations rather than autoregressive prediction targets; supervision is applied only to the reasoning tokens, camera actions, and final response.

The overall SFT objective is
\begin{equation}
	\mathcal{L}_{\mathrm{SFT}}
	=
	\lambda_r \mathcal{L}_{\mathrm{reason}}
	+
	\lambda_a \mathcal{L}_{\mathrm{action}}
	+
	\lambda_f \mathcal{L}_{\mathrm{final}},
	\label{eq:sft_objective}
\end{equation}
where
\begin{equation}
	\mathcal{L}_{\mathrm{reason}}
	=
	-
	\sum_{t=1}^{T}
	\sum_{k \in R_t^{*}}
	\log
	p_{\theta}
	\left(
	r_{t,k}^{*}
	\mid
	C_t,
	r_{t,<k}^{*}
	\right),
	\label{eq:reason_loss}
\end{equation}
and
\begin{equation}
	\mathcal{L}_{\mathrm{action}}
	=
	-
	\sum_{t=1}^{T}
	\log
	p_{\theta}
	\left(
	A_t^{*}
	\mid
	C_t,R_t^{*}
	\right).
	\label{eq:action_loss}
\end{equation}
Here, $C_t$ contains the initial multimodal context together with all preceding reasoning steps, actions, and returned world-model latents. The final-response loss $\mathcal{L}_{\mathrm{final}}$ is the standard autoregressive loss over the final justification and answer. We use separate terms because reasoning, camera actions, and final responses constitute distinct supervised outputs, whereas the returned world-model latents are conditioning observations rather than prediction targets [\citenum{zhou2026autovla,yang2026context}].

In practice, such interleaved reasoning--action--latent trajectories are not directly available in existing spatial-reasoning datasets. We therefore construct synthetic reasoning trajectories using a teacher model.%

During this stage, we jointly optimize the latent adapter $\mathcal{P}_{w}$ and lightweight LoRA modules inserted into the VLM, while keeping the pretrained VLM backbone and world model frozen. Unlike latent alignment, which teaches the model to interpret world-model features, trajectory SFT teaches it how to use those features within an iterative spatial reasoning process.

\subsection{Process Verifiable RL}
\label{sec:rl}

While supervised fine-tuning provides a strong initialization for spatial reasoning and action generation, it primarily teaches the model to imitate teacher-provided trajectories rather than directly optimize task performance. We therefore further train WorldMind with reinforcement learning, allowing the model to explore its own reasoning and camera-action trajectories and optimize them according to task-level rewards. This stage encourages the model to discover more effective spatial transformations, improve final-answer accuracy, and avoid unnecessary simulation steps.

At step $t$, the policy samples $(R_t,A_t)\sim\pi_\theta(\cdot\mid C_t)$. The action $A_t$ is executed by the world model to produce $\widetilde{Z}_t=\mathcal{W}(S_{t-1},A_t)$, which is projected as $\widetilde{H}_t^{w}=\mathcal{P}_w(\widetilde{Z}_t)$ and appended to the context for the next reasoning step. This yields an interactive rollout
\begin{equation}
	S_0
	\xrightarrow{\pi_\theta}
	A_1
	\xrightarrow{\mathcal{W}}
	S_1
	\xrightarrow{\pi_\theta}
	\cdots
	\xrightarrow{\mathcal{W}}
	S_T
	\rightarrow Y.
	\label{eq:rl_rollout}
\end{equation}

A central challenge is assigning credit not only to the final answer, but also to the intermediate reasoning and simulation steps that produce it. Conventional outcome rewards, such as answer correctness and format compliance, provide little guidance on whether a camera transformation is useful or whether the resulting world-model state contributes to subsequent reasoning [\citenum{li2026imagination,guo2026beyond}]. We therefore augment it with a step-wise reward.

Conceptually, we impose three requirements on each reasoning step. First, the newly acquired evidence should improve the model's ability to answer the question compared with reasoning from the previous context alone. We therefore define the \textbf{progress reward} [\citenum{setlur2025rewarding}] as
\begin{equation}
	r_t^{\mathrm{prog}}
	=
	\Delta V_t
	=
	V(C_{t+1}) - V(C_t),
	\label{eq:progress_reward}
\end{equation}
where $V(C_t)$ denotes the estimated probability of reaching the correct answer from context $C_t$. Second, the selected camera action should correctly realize the intended spatial instruction. We capture this with the \textbf{action validity} $c_t^{\mathrm{action}}$, which measures if the generated action $A^j$ is consistent with the textual instruction $R^j$ in this reasoning step. Third, because the synthesis capability of a world model is inherently bounded (e.g. it may become unreliable when extrapolating far beyond the region covered by the observed views) we introduce the \textbf{state validity} $c_t^{\mathrm{state}}$, which measures whether the resulting latent state remains within the reliable operating range of the world model.

Rather than adding these three terms independently, we use action and state validity as multiplicative gates [\citenum{pan2020trust,park2026know}] on progress:
\begin{equation}
	r_t^{\mathrm{step}}
	=
	c_t^{\mathrm{action}}
	c_t^{\mathrm{state}}
	\Delta V_t.
	\label{eq:step_reward}
\end{equation}
This design prevents the policy from earning process reward merely by selecting easy-to-execute actions or producing plausible-looking states. High reward is obtained only when a step simultaneously makes useful progress, correctly executes the intended transformation, and produces a reliable world-model state. It therefore discourages degenerate behaviors such as repeatedly taking safe but uninformative actions or exploiting hallucinated latent states that spuriously increase answer confidence.

The overall trajectory reward is then
\begin{equation}
	R(\tau)
	=
	\lambda_{\mathrm{ans}} r_{\mathrm{ans}}
	+
	\lambda_{\mathrm{fmt}} r_{\mathrm{fmt}}
	+
	\sum_{t=1}^{T} r_t^{\mathrm{step}}
	-
	\lambda_c T,
	\label{eq:trajectory_reward}
\end{equation}
where $r_{\mathrm{ans}}$ evaluates final-answer correctness, $r_{\mathrm{fmt}}$ enforces the required output format, and $\lambda_c T$ penalizes unnecessary world-model calls.

In practice, we compute $c_t^{\mathrm{latent}}$ by measuring the similarity between the world-model latent generated after executing $A_t$ and the latent representation of the corresponding ground-truth target view. For synthetic 3D scenes, we render the ground-truth observation $O_t^{\mathrm{GT}}$ at the same target camera pose and extract its latent representation $H_t^{\mathrm{GT}}$ using the same world-model layer and diffusion timestep as the generated latent $H_t^w$. Since the two representations lie in the same latent space and share the same spatial organization, we measure their consistency using the average cosine similarity between spatially corresponding latent tokens. Specifically, let $\widetilde{H}_t^w=\{h_{t,i}^w\}_{i=1}^{N}$ and $\widetilde{H}_t^{\mathrm{GT}}=\{h_{t,i}^{\mathrm{GT}}\}_{i=1}^{N}$. We define
\begin{equation}
	c_t^{\mathrm{state}}
	=
	\frac{1}{N}
	\sum_{i=1}^{N}
	\frac{
		1+\cos\left(h_{t,i}^{w},h_{t,i}^{\mathrm{GT}}\right)
	}{2},
	\label{eq:latent_reliability}
\end{equation}
where the cosine similarity is rescaled from $[-1,1]$ to $[0,1]$. A high $c_t^{\mathrm{latent}}$ indicates that the generated latent remains consistent with the ground-truth scene state at the requested viewpoint, whereas a low value suggests that the transformation exceeds the reliable operating range of the world model and may introduce hallucinated or spatially inconsistent content.

\section{Experiments}
\label{sec:experiments}

\paragraph{Evaluation Benchmarks.}
We evaluate WorldMind on multiple benchmarks that include complementary aspects of visual spatial reasoning, such as metric understanding, relative spatial relations, perspective taking, multi-view reasoning, and navigation-oriented reasoning. These aspects widely cover video-based spatial understanding, viewpoint transformation and compositional reasoning, allowing us to assess where latent world simulation provides the largest benefit.
\vspace{-0.1in}
\begin{itemize}
	\item \textbf{VSI-Bench}~[\citenum{zhu2026thinking}] evaluates video-based spatial reasoning in indoor environments across diverse task categories, including object counting, absolute and relative distance, object and room size, relative direction, route planning, and appearance order. Following the official evaluation protocol, we report Accuracy for multiple-choice questions and Mean Relative Accuracy (MRA) for numerical questions.
	\vspace{-0.05in}
	\item \textbf{MindCube}~[\citenum{yin2025spatial}] evaluates spatial mental modeling from limited multi-view observations. It contains three settings (Rotation, Among, and Around )that test whether a model can construct and manipulate an internal representation of a scene from only a few views. We report accuracy for each setting together with the overall score.
	\vspace{-0.05in}
	\item \textbf{MMSI-Bench}~[\citenum{yang2026mmsi}] provides a broader evaluation of multimodal spatial intelligence across 11 task categories, covering camera--camera, camera--object, camera--region, object--object, object--region, and region--region relations, as well as measurement, appearance, camera motion, object motion, and multi-step reasoning. We report per-category accuracy together with the overall benchmark score.
\end{itemize}
\vspace{-0.1in}

\vspace{-0.05in}
\paragraph{Evaluation Baselines.}
We compare WorldMind with three groups of competitive baselines:
\vspace{-0.1in}
\begin{itemize}
	\item \textbf{General-purpose VLMs.}
	We use Qwen3-VL-8B-Instruct [\citenum{bai2025qwen3vltechnicalreport}], LLaVA-OneVision-7B [\citenum{li2024llava}], and LLaVA-Video-7B [\citenum{zhang2024llava}] as general-purpose multimodal baselines. They directly reason over the observed visual inputs without explicit 3D representations or world model simulation.
	\vspace{-0.05in}
	\item \textbf{Spatial-Aware VLMs.}
	We compare WorldMind against VLM-3R-7B [\citenum{fan2026vlm}] and Cambrian-S-7B [\citenum{yang2026cambrian}], which enhance general-purpose VLMs with explicit geometric or spatial representations. These baselines test whether directly injecting 3D-aware features is sufficient for spatial reasoning, without actively transforming the represented scene.
	\vspace{-0.05in}
	\item \textbf{VLMs with world-model simulation.}
	We further compare with SpatialDreamer \citenum{cao2026spatialdreamer}, ASTRA \citenum{zhu2026thinking} and AVIC \citenum{yu2026and}, which couple VLMs with world models to acquire additional observations through active visual imagination. Unlike WorldMind, these methods primarily return synthesized RGB pixel views to the VLM, providing the closest comparison against WorldMind's reasoning directly from latent world simulation.
\end{itemize}
\vspace{-0.1in}

\begin{table}
	\centering
	{\fontsize{8}{10}\selectfont
        \setlength{\tabcolsep}{5pt}
		\begin{tabular}{lccccccccc}
			\toprule
			\textbf{Method} & \rotatebox{55}{\textbf{Obj. Count}} & \rotatebox{55}{\textbf{Abs. Dist.}} & \rotatebox{55}{\textbf{Rel. Dist.}} & \rotatebox{55}{\textbf{Obj. Size}} & \rotatebox{55}{\textbf{Room Size}} & \rotatebox{55}{\textbf{Rel. Dir.}} & \rotatebox{55}{\textbf{Route Plan.}} & \rotatebox{55}{\textbf{App. Order}} & \rotatebox{55}{\textbf{Avg.}} \\
			\midrule
			LLaVA-Video-7B & 48.5 & 14.0 & 43.5 & 47.8 & 24.2 & 42.4 & 34.0 & 30.6 & 35.6 \\
			LLaVA-OneVision-7B & 47.7 & 20.2 & 42.5 & 47.4 & 12.3 & 35.2 & 29.4 & 24.4 & 32.4 \\
			Qwen3-VL-8B-Instruct & 66.7 & 38.9 & 59.0 & 74.1 & 53.5 & 65.4 & 32.0 & 69.4 & 57.4 \\
			VLM-3R-7B & 70.2 & 49.4 & 65.4 & 69.2 & 67.1 & 80.5 & 45.4 & 40.1 & 60.9 \\
			Cambrian-S-7B & 73.2 & 50.5 & 71.1 & 74.9 & \textbf{72.2} & 76.2 & 41.8 & \textbf{80.1} & 67.5 \\
			SpatialDreamer & 71.1 & 48.1 & 67.8 & 67.0 & 65.9 & 86.4 & 47.5 & 37.2 & 62.2 \\
			AVIC & 72.7 & 49.8 & 65.3 & 74.8 & 57.9 & 81.2 & 37.9 & 63.8 & 66.6 \\
			\midrule
			\textbf{WorldMind (SFT)} & \textbf{73.2} & \textbf{59.5} & \textbf{72.9} & 74.9 & 65.3 & \textbf{86.7} & \textbf{50.6} & 76.2 & \textbf{67.9} \\
			\textbf{WorldMind (SFT+RL)} & 71.4 & \textbf{62.3} & \textbf{73.2} & \textbf{75.7} & 66.8 & \textbf{87.1} & \textbf{55.4} & 77.1 & \textbf{71.1} \\
			\bottomrule
	\end{tabular}
	}

	\caption{Comparison of task accuracy with the existing methods on the VSI-Bench benchmark. ``Obj.'' indicates Objects, ``Abs. Dist.'' indicates Absolute Distance, ``Rel. Dist.'' indicates Relative Distance, ``Rel. Dir.'' indicates Relative Direction, ``Route Plan.'' indicates Route Planning, and ``App. Order'' indicates Appearance Order.}
	\vspace{-0.1in}
	\label{tab:vsi_results}
\end{table}

\begin{table*}[t]
	\centering
	{\fontsize{8}{10}\selectfont
		\begin{tabular}{lccccc|cccc}
			\toprule
			\textbf{Model}
			& \multicolumn{5}{c|}{\textbf{MMSI-Bench}}
			& \multicolumn{4}{c}{\textbf{MindCube-Tiny}} \\
			\cmidrule(lr){2-6}\cmidrule(lr){7-10}
			& \textbf{PR.} & \textbf{Attr.} & \textbf{Mot.} & \textbf{MSR} & \textbf{All}
			& \textbf{Rot.} & \textbf{Ard.} & \textbf{Amg.} & \textbf{All} \\
			\midrule

			LLaVA-NeXT-Video-7B
			& 23.5 & 30.0 & 21.3 & 27.8 & 25.7
			& 35.7 & 30.1 & 43.6 & 36.5 \\

			LLaVA-OneVision-7B
			& 28.0 & 27.7 & 26.7 & 11.6 & 23.5
			& 36.5 & 44.1 & 48.4 & 43.0 \\

			Qwen3-VL-8B-Instruct
			& 33.7 & 26.2 & 28.0 & 29.8 & 29.4
			& 53.6 & 38.0 & 31.1 & 40.9 \\

			VLM-3R-7B
			& 33.0 & 31.5 & 32.1 & 30.3 & 31.7
			& 36.7 & 24.5 & 44.2 & 35.1 \\

			Cambrian-S-7B
			& 29.5 & 28.4 & 21.4 & 24.2 & 25.9
			& 33.0 & 39.2 & 39.0 & 37.1 \\

			Astra (Qwen3-VL-8B-Instruct)
			& 42.3 & 41.0 & 32.1 & 33.6 & 37.3
			& 60.1 & 43.5 & 36.8 & 46.8 \\

			AVIC
			& 31.1 & \textbf{45.4} & 32.0 & 28.2 & 34.2
			& 36.8 & 40.5 & 38.4 & 38.6 \\

			\midrule

			\textbf{WorldMind (SFT)} & \textbf{44.5} & 44.9 & \textbf{35.6} & \textbf{37.7} & 40.7 & \textbf{65.2} & \textbf{46.1} & 48.1 & 53.1 \\

			\textbf{WorldMind (SFT+RL)} & \textbf{47.7} & 42.8 & \textbf{40.9} & \textbf{42.2} & \textbf{43.4} & \textbf{71.6} & \textbf{50.3} & \textbf{55.8} & \textbf{59.2} \\

			\bottomrule
	\end{tabular}
	}

	\caption{Comparison of task success rate with the existing methods on the MMSI-Bench and MindCube benchmarks. For MMSI-Bench, ``PR.'' indicates Positional Relationship, ``Attr.'' indicates Attribute, ``Mot.'' indicates Motion, and ``MSR'' indicates Multi-Step Reasoning. For MindCube, ``Rot.'' indicates Rotation, ``Ard.'' indicates Around, and ``Amg.'' indicates Among. ``All'' denotes the unweighted mean across the reported task categories. Higher is better ($\uparrow$).}
	\vspace{-0.2in}
	\label{tab:mmsi_mind}
\end{table*}

\vspace{-0.1in}
\paragraph{Training Data.}
WorldMind is trained on a mixed collection of data from multiple sources. Besides using multiple real-world datasets such as VSI-590K \citenum{yang2026cambrian}, SPAR-234K \citenum{zhang2026flatland}, SpatialMosaic \citenum{lee2025spatialmosaic}, and SAT-Real \citenum{ray2024sat}, we additionally incorporate synthetic datasets, including SAT-Synthesized \citenum{ray2024sat} in AI2-THOR \citenum{kolve2017ai2} and Infinibench \citenum{raistrick2024infinigen,wang2026infinibench} scenes, for trajectory construction and reinforcement learning, to provide ground-truth novel views and scene states at the target camera poses in world simulation. For latent alignment, we sample a total of 40,000 examples across the different real-world data sources. We then construct 10,000 synthetic reasoning trajectories for supervised fine-tuning using GPT-5.6 Sol as the teacher model. For reinforcement learning, we sample 5,000 questions and allow the SFT-initialized policy to generate its own on-policy reasoning and simulation trajectories. No samples from the evaluation splits of the benchmarks reported in the main results are used in training.%

\vspace{-0.1in}
\paragraph{Implementation.}%
We use SVC [\citenum{zhou2025stable}] as the frozen world model. The latent features of world model are projected into the VLM embedding space using a lightweight Transformer adapter, while LoRA modules are added to the VLM during SFT and reinforcement learning. We extract candidate features from four world-model layers and four denoising timesteps, and select the top-$8$ layer-timestep pairs offline using feature ablation. The adapter is first aligned for one epoch, followed by 3 epochs of simulated-trajectory SFT. We then perform GRPO initialized from the SFT checkpoint with 8 rollouts per question. All experiments use BF16 precision.

\subsection{Main Results}
\label{sec:main_results}
Tables~\ref{tab:vsi_results} and \ref{tab:mmsi_mind} summarize the main experiment results on the three spatial benchmarks. Overall, WorldMind consistently improves over the underlying VLM backbone and remains competitive with or outperforms specialized spatial-reasoning baselines. In particular, WorldMind (SFT+RL) achieves an average improvement of 16.0\% over the base VLM and 8.3\% over the strongest RGB-based world-model baseline across the evaluated benchmarks.

The gains are especially pronounced on tasks that require reasoning under transformed viewpoints, multi-step spatial relations, or camera/object motion. This trend is consistent with the design of WorldMind, which explicitly acquires additional spatial evidence through latent world simulation rather than relying only on the original observations. By contrast, the improvements are relatively smaller on tasks that can be solved directly from the visible scene, such as Object Counting and Object Size, suggesting that the main benefit of WorldMind comes from reasoning about spatial states that are not explicitly observed.

Comparison with the two WorldMind variants further shows that simulation-trajectory SFT already provides substantial gains by teaching the model how to interact with the world model and interpret latent states, while process-verifiable reinforcement learning provides an additional improvement of 4.0\%. This indicates that optimizing the policy on its own trajectories helps it select more useful camera transformations and avoid unreliable or unnecessary simulations.

\subsection{Ablation Studies}
\label{sec:ablation}

\begin{wraptable}{r}{0.59\linewidth}
	\vspace{-10pt}
	\centering
	{\fontsize{8}{10}\selectfont\setlength{\tabcolsep}{5.8pt}
	\begin{tabular}{lccc}
		\toprule
		\textbf{Method}
		& \textbf{VSI-Bench}
		& \textbf{MMSI-Bench}
		& \textbf{Mindcube} \\
		\midrule
		Base VLM
		& 57.4 & 29.4 & 40.9 \\
		+ Linguistic Scaling
		& 55.9 & 29.7 & 40.7 \\
		+ RGB Simulation
		& 63.4 & 33.6 & 48.4 \\
		\textbf{+ Latent Simulation}
& \textbf{65.2} & \textbf{36.2} & \textbf{48.5} \\
		\bottomrule
	\end{tabular}
}

	\caption{Ablation of the reasoning interface.}
	\vspace{-0.1in}
	\label{tab:ablation_interface}
\end{wraptable}

\paragraph{Effect of latent world simulation.} We first isolate the contribution of latent world simulation by comparing the base VLM using only observed RGB frames, additional linguistic test-time reasoning, RGB-based world simulation, and WorldMind with latent world simulation. As shown in Tab.~\ref{tab:ablation_interface}, linguistic scaling provides only limited and inconsistent benefits. In contrast, RGB-based world simulation substantially improves performance across all three benchmarks. Replacing RGB outputs with directly exposed world-model latents yields further gains, improving the average performance from 48.5 to 50.0 and supporting the benefit of avoiding the latent-to-pixel-to-latent reasoning bottleneck. These results are from a small-scale experiment and therefore differ from the full-scale results in the main comparison.

\begin{wraptable}{r}{0.4\linewidth}
	\vspace{-10pt}
	\centering

	{\fontsize{8}{10}\selectfont
	\begin{tabular}{lcc}
		\toprule
		\textbf{Selection}
		& $\mathbf{K}$
		& \textbf{Task success rate} \\
		\midrule
		Single feature
		& 1 & 59.8 \\
		Random
		& 4 & 57.8 \\
		Full feature bank
		& All & 60.3 \\
		\textbf{Ours}
		& 8 & \textbf{60.4} \\
		\bottomrule
	\end{tabular}
	}

	\caption{Ablation of latent feature selection on the VSI-Bench benchmark}
	\label{tab:ablation_selection}

	\vspace{-8pt}
\end{wraptable}

\paragraph{Latent feature selection.}
Tab.~\ref{tab:ablation_selection} examines which world-model representations are most useful for spatial reasoning. We conduct a small-scale experiment on VSI-Bench using partial latent alignment and SFT, without reinforcement learning. We compare a single layer--timestep feature, randomly selected features, the complete feature bank, and our offline selected subset. Our selected features achieve \textbf{60.4}, compared with \textbf{60.3} for the full feature bank, indicating that a compact subset of informative representations is sufficient while avoiding redundant or less task-relevant features. In contrast, randomly selecting four features degrades performance to 57.8, further highlighting the importance of selecting spatially informative layer--timestep representations rather than indiscriminately using the complete feature bank.

\vspace{4pt}

\begin{wraptable}{r}{0.6\linewidth}
	\vspace{-10pt}
	\centering
		{\fontsize{8}{10}\selectfont
	\begin{tabular}{lcccc}
		\toprule
		\textbf{Reward}
		& $\Delta V$
		& $c^{\mathrm{action}}$
		& $c^{\mathrm{state}}$
		& \textbf{Success rate} \\
		\midrule
		Final answer only
		& -- & -- & -- & 68.1 \\
		+ Progress
		& \checkmark & -- & -- & 68.3 \\
		+ Action validity
		& \checkmark & \checkmark & -- & 68.9 \\
		+ Latent validity
		& \checkmark & -- & \checkmark & 68.0 \\
		\textbf{Full}
		& \checkmark & \checkmark & \checkmark & \textbf{70.2} \\
		\bottomrule
	\end{tabular}
	}

	\caption{Ablation of the process reward on VSI-Bench}
	\vspace{-0.1in}
	\label{tab:ablation_reward}
\end{wraptable}

\paragraph{Reward design.}
Tab.~\ref{tab:ablation_reward} evaluates the components of the process-verifiable reward. Starting from terminal answer supervision alone, we progressively introduce rollout-derived progress $\Delta V_t$, action validity $c_t^{\mathrm{action}}$, and latent validity $c_t^{\mathrm{latent}}$. Progress supervision provides a modest improvement from \textbf{68.1} to \textbf{68.3}, while additionally incorporating action validity further improves performance to \textbf{68.9}. Latent validity alone does not improve over progress supervision, yielding 68.0; however, combining both validity terms with progress substantially increases performance to \textbf{70.2}. The full formulation therefore performs best, indicating that action and latent validity provide complementary signals and that useful intermediate progress should be credited only when the intended transformation is successfully realized and the resulting world-model latent remains reliable.

\Needspace{7\baselineskip}
\section{Conclusion}
We introduce \textbf{WorldMind}, a framework for test-time spatial reasoning via latent world simulation. By directly conditioning the VLM on world-model latents, it enables iterative spatial transformation without costly RGB conversion. Experiments show consistent gains, particularly on viewpoint-dependent and multi-step reasoning tasks.

\setcitestyle{numbers}
\bibliographystyle{abbrvnat}
\begingroup
\setlength{\bibsep}{1pt}
\bibliography{references}
\endgroup

\end{document}